\documentclass[runningheads]{llncs}
\usepackage[T1]{fontenc}
\usepackage{graphicx}
\usepackage{booktabs}
\usepackage[misc]{ifsym}

\usepackage{booktabs}
\usepackage{algorithm}
\usepackage{algorithmic}
\usepackage{multirow}
\usepackage{makecell}
\usepackage{subcaption}

\usepackage{amsfonts}
\usepackage{amsmath}
\usepackage{graphicx}
\usepackage{arydshln}
\usepackage[skip=6pt]{caption}

\usepackage{mwe}

\begin{document}

\title{Joint Structural Pruning and Mixed-Precision Quantization for LLM Compression}

\titlerunning{A LLM compression method for jointly pruning and mixed precision quantization}
\author{Hoang-Loc La\inst{1}\orcidID{0009-0005-5453-7836} \and
Truong-Thanh Le\inst{1,2}\orcidID{0009-0000-3515-8244} \and
Amir Taherkordi\inst{2} \and 
Phuong Hoai Ha\inst{1}\orcidID{0000-0001-8366-5590}}

\institute{UiT The Arctic University of Norway \and
University of Oslo, Norway \\
\email{\{hoang.l.la,phuong.hoai.ha\}@uit.no \\\{truongl,amirhost\}@ifi.uio.no}}
\maketitle              

\begin{abstract}
Recently, the efficiency of Large Language Models (LLMs) deployment has become a critical concern in practical applications. While post-training quantization (PTQ) and structural pruning are established techniques for reducing memory footprint and inference latency, most existing PTQ approaches optimize quantization errors on a per-layer basis, overlooking how errors accumulate and propagate through the network, often resulting in suboptimal solutions. Traditional pipelines also tend to apply pruning and quantization in isolation or sequentially, further compounding sub-optimality. We introduce a novel end-to-end framework that addresses these limitations in two key ways. First, we propose a novel mixed-precision PTQ strategy that directly minimizes global error propagation across the entire model, rather than isolating layer-wise errors. Building on this, we develop a novel joint optimization approach that simultaneously learns structural pruning decisions and mixed-precision quantization policies within a unified search space. Extensive experiments show that, at ultra-low precisions (1–3 bits), our quantization method reduces WikiText perplexity by up to 21\% compared to state-of-the-art (SoTA) weight-activation quantization baselines. Against leading weight-only quantization methods, it achieves up to 59\% and 85\% lower perplexity on WikiText and C4, respectively. Compared to the SoTA joint pruning-and-quantization techniques, our proposed method delivers superior perplexity and reasoning performance at ultra-low bits. Furthermore, in mainstream mixed-precision settings (e.g., 4-bit/8-bit), our compressed models remain highly competitive in terms of perplexity on WikiText and zero-shot reasoning accuracy, while delivering up to $2\times$ faster prefill, $6.5\times$ peak memory reduction during decoding (vs. FP16), up to 30\% faster inference, and an additional 10\% memory savings compared to models compressed by SoTA methods.

\end{abstract}

\section{Introduction}
\par The efficiency of large language models (LLMs) deployment has become a major research focus in recent years. To enable deployment on resource-constrained devices, prior works have extensively investigated model compression for pre-trained LLMs. Pruning and quantization are the two most common techniques, as they reduce memory footprint and computational requirements while preserving reasoning performance for efficient edge inference.
\par Pruning eliminates redundant parameters to reduce resource usage. However, compared to quantization at the same high compression ratio (e.g., 75\% ), pruned models often exhibit inferior reasoning performance~\cite{kuzmin2023pruning}. Quantization maps weights (and optionally activations) to lower-bit representations and typically preserves better performance for a given compression ratio. It encompasses quantization-aware training (QAT), which requires fine-tuning, and post-training quantization (PTQ), which operates directly on pre-trained weights without retraining. This work focuses on PTQ, as it avoids the costly retraining or fine-tuning steps associated with QAT.
\par Recent mixed-precision PTQ methods ~\cite{zhao2024atom,zhao2025ptq1,huang2024slim,huang2024billm} for ultra-low-bit LLMs typically partition linear-layer weights into salient and non-salient channels. Non-salient channels are quantized to very low precisions (e.g., 1-4 bits), while salient ones retain higher precision (e.g., 8 bits). These methods generally rely on layer-wise greedy search guided by local quantization loss, which overlooks global error propagation and applies uniform saliency criteria across layers despite varying inter-layer sensitivities. This often results in suboptimal compression~\cite{arai2025quantization}. Our work introduces a novel mixed-precision PTQ approach to address these limitations.
\par To further improve reasoning and generation abilities under aggressive compression, recent studies also explore joint pruning and quantization, showing the two are non-orthogonal both theoretically and empirically~\cite{harmaeffective}. Methods like SparseGPT+GPTQ~\cite{frantar2023sparsegpt} and OBR~\cite{guo2025optimal} support joint unstructured/semi-structured compression with error compensation, but offer limited speedup compared to structured methods. We introduce a joint structured pruning and mixed-precision PTQ framework that combines our novel quantization approach with DISP-LLM~\cite{gao2024disp}, enabling co-optimization over the joint search space.
\par Our main contributions are as follows:
\begin{itemize}
    \item We introduce a novel mixed-precision post-training quantization (PTQ) framework that reformulates bit-width allocation as a binary mask optimization problem. A hypernetwork, trained directly on the end-to-end task loss, learns and optimizes these binary masks. Unlike prior methods that rely on fixed, hand-crafted thresholds to separate salient from non-salient weights, our approach dynamically identifies the most important (salient) weights per linear layer using the global end-to-end loss. This adaptive, loss-driven strategy overcomes the local greedy nature and suboptimality of earlier techniques. Compared to state-of-the-art (SoTA) weight-activation PTQ baselines, our method reduces WikiText-2 perplexity by up to 21\% and improves average zero-shot accuracy across six reasoning benchmarks by up to 4.5\%. Against SoTA weight-only PTQ baselines, it achieves up to 59\% and 85\% lower perplexity on WikiText-2 and C4, respectively, along with up to 5.4\% higher average zero-shot accuracy on the reasoning tasks.
    \item An integrated structured pruning and mixed-precision quantization framework, named \textbf{train-once-get-all (TOGA)}, that delivers SoTA perplexity and zero-shot reasoning performance across diverse tasks.
    \item Custom CUDA kernels for efficient mixed-precision matrix multiplication (e.g., W4A4 + W8A8). Leveraging these kernels, our compressed models achieve up to $2\times$ prefill speedup and $6.5\times$ peak memory reduction compared to FP16, while outperforming the strongest prior 2:4 semi-structured sparsity techniques by 30\% faster prefill and 10\% peak memory savings.
\end{itemize}
\section{Related Work and Our Advancements beyond SoTA}
\subsection{Mixed-precision Post-Training Quantization techniques}
\par Recent mixed-precision quantization (MPQ) methods assign higher bit-widths to salient (sensitive) weight channels according to their quantization error sensitivity, typically estimated from local layer-level information such as Hessian information, gradients, or activation statistics~\cite{huang2024slim,huang2024billm,zhao2024atom,zhao2025ptq1}. Weight-only MPQ has substantially improved memory efficiency in memory-bound inference scenarios, while activations remain at 16-bit precision. For instance, PTQ-1.61~\cite{zhao2025ptq1} and BiLLM~\cite{huang2024billm} use per-layer Hessian-based metrics to distinguish salient from non-salient weights. Slim-LLM~\cite{huang2024slim} applies a local greedy search within each linear layer to determine optimal bit-width allocations for its weight matrix. 

\par Weight-activation MPQ directly addresses activation outliers for greater efficiency gains. Atom~\cite{zhao2024atom} preserves critical activations by identifying them based on their magnitude, reordering them (along with corresponding weight channels) to the end of the matrices, then quantizing salient channels to higher precision (8-bit) while applying lower precision (3- or 4-bit) to non-salient ones. In another approach, ResQ~\cite{saxena2024resq} employs principal component analysis (PCA) to separate sensitive and non-sensitive components, applying W4A4 quantization to non-sensitive weight-activation channels and W8A8 to the rest.
\par \textbf{Advancement beyond SoTA:} These methods share two key limitations. First, they typically rely on fixed, uniform saliency thresholds to identify sensitive weights, thereby overlooking differences in layer-wise sensitivity across models and architectures. Second, they allocate bits using only local, layer-level sensitivity metrics. As a result, they fail to account for the cumulative quantization error that accumulates as signals propagate through the entire network \cite{arai2025quantization}. Our approach addresses both issues by optimizing binary masks to identify salient weights across the entire model architecture, thereby enabling flexible, non-uniform bit allocation across linear layers, while directly optimizing the global language modeling loss.

\subsection{Pruning+Quantization methods}
\par Harma et al.~\cite{harmaeffective} demonstrate that pruning and quantization are non-orthogonal, and that the order of applying them significantly impacts performance. In LLMs, a prune-then-quantize sequence consistently yields lower perplexity than quantize-then-prune or applying either technique in isolation. Similarly, in vision models, Kuzmin et al.~\cite{kuzmin2023pruning} show that quantized models generally preserve higher accuracy than pruned models at equivalent compression ratios. Their findings further reveal that combining mild pruning with higher-precision quantization produces a superior accuracy compared to aggressive low-bit quantization alone.
\par  SparseGPT~\cite{frantar2023sparsegpt} builds on the Optimal Brain Surgeon (OBS) framework~\cite{hassibi1993optimal} to perform unstructured pruning and quantization of LLMs. The authors show that applying both techniques jointly by combining SparseGPT with GPTQ~\cite{frantar2022gptq} outperforms using either method alone. More recently, OBR~\cite{guo2025optimal} introduces an explicit error-compensation step between pruning and quantization and uses the OBS principles to better align their conflicting effects on weight distributions.
\par \textbf{Advancement beyond the SoTA:} Both SparseGPT and OBR rely on unstructured or semi-structured sparsity patterns. While unstructured sparsity offers strong theoretical compression, it provides limited practical inference speedups on standard GPU hardware~\cite{hoefler2021sparsity}. On the other hand, semi-structured sparsity can deliver practical speedups, but it is constrained by current hardware: NVIDIA GPUs primarily support only the 2:4 pattern (50\% sparsity with a specific 2-out-of-4 non-zero layout). When targeting other sparsity ratios (e.g., 40\% or 60\%), these methods typically fall back to fully unstructured pruning, which remains highly inefficient on existing hardware and yields little to no real speedup. In contrast, to the best of our knowledge, this work introduces the first joint \emph{structured pruning} and \emph{mixed-precision quantization} framework for large language models. By producing hardware-friendly, dense-compatible matrices, our approach enables efficient execution on standard GPUs while consistently achieving superior accuracy--efficiency trade-offs compared to prior joint unstructured or semi-structured methods. Moreover, the structurally pruned models produced by our method achieve up to 30\% faster inference, 10\% lower peak memory usage with better perplexity and reasoning performance than 2:4 semi-structured pruned models from previous works. Please see Section~\ref{sect:exp} for detailed experimental results.

\subsection{Structural Pruning for LLM with Binary Masks}
\label{subsect:hypernetwork}
\par \textbf{Transformers:} LLMs are predominantly based on the decoder-only transformer architecture~\cite{vaswani2017attention}. Specifically, each transformer block comprises two main submodules: multi-head self-attention (MHA) and a feed-forward network (FFN), each followed by a residual connection and layer normalization. Hence, given input $X$ to a transformer block, the core computations are described as:
\begin{align}
    \text{Attention}(X) &= \text{MHA}(XW_q, XW_k, XW_v)W_o, \\
    \text{MLP}(X) &= \bigl(\sigma(XW_{\text{gate}}) \odot (XW_{\text{up}})\bigr) W_{\text{down}},
\end{align}
where MHA captures position-wise dependencies using multiple attention heads, each defined by linear projections for queries ($W_q$), keys ($W_k$), values ($W_v$), and output ($W_o$). The FFN (also called MLP) applies gate ($W_{\text{gate}}$), up-projection ($W_{\text{up}}$), and down-projection ($W_{\text{down}}$) matrices, with non-linearity $\sigma$ after the gate. 
\par \textbf{Binary Masks:} Our proposed method draws inspiration from DISP-LLM~\cite{gao2024disp}, which frames structured pruning as a learnable binary mask optimization problem. Let $L$ denote the total number of binary masks across the model, and let $\mathcal{P} = \{P_i\}_{i=1}^L$ be the set of learnable binary masks, where each $P_i \in \{0,1\}^{d_i}$ indicates which channels (input or output) of the corresponding weight matrix are retained ($1$) or pruned ($0$). For a linear layer with full-precision weight matrix $W \in \mathbb{R}^{d_{\text{out}} \times d_{\text{in}}}$, structured pruning is applied using an input mask $P_{\text{in}} \in \{0,1\}^{d_{\text{in}}}$ and an output mask $P_{\text{out}} \in \{0,1\}^{d_{\text{out}}}$. The pruned weight matrix is obtained as:
\begin{align}\label{eq:prune}
    F_{\text{prune}}(W, P_{\text{in}}, P_{\text{out}}) = \operatorname{diag}(P_{\text{in}}) \, W \, \operatorname{diag}(P_{\text{out}}) = P_{\text{in}}^T W P_{\text{out}}.
\end{align}

\par Applying this to the attention and feed-forward modules yields:
\begin{align}
    \text{Attention}(X) &= \text{MHA}(X P_1, X P_1, X P_1) (W_o P_2), \\
    \text{MLP}(X) &= \bigl(\sigma(X P_3 W_{\text{gate}}) \odot (X P_3 W_{\text{up}})\bigr) (P_4^T W_{\text{down}} P_5),
\end{align}
where $\{P_i\}_{i=1}^5$ are the pruning masks for linear layers in a transformer block.
\par Following DISP-LLM, in this work, we only prune the input and output dimensions of attention modules and the input, intermediate, and output dimensions of the MLP module, while keeping the number of attention heads and head dimension fixed.
\par \textbf{Search for optimal binary masks:} Let $S$ denote a set of $L$ binary vectors $s_l$ that govern the joint pruning and quantization of the model, which consists of $L$ linear layers. While searching for the optimal configuration $S^*$ can be resolved with computationally intensive techniques, namely, evolutionary algorithms~\cite{tang2025darwinlm} and reinforcement learning, motivated by DISP-LLM, we advocate for a more efficient hypernetwork-based approach.  
To guide the hypernetwork towards configurations that satisfy a desired budget $b$, let $B(S)$ be a differentiable function that estimates the expected budget (e.g., effective sparsity, bit-width-averaged memory usage, or memory savings) induced by configuration $S$. The budget regularization term can be defined as follows.
\begin{align}
    R(b, B(S)) =  \log(\frac{\max(b, B(S))}{\min(b, B(S))})
\end{align}
\par This term regularizes the expected budget $B(S)$ to be match the target $b$. Following DISP-LLM~\cite{gao2024disp}, the hypernetwork is trained using a loss function that combines the standard cross-entropy language modeling objective $L_{\text{CE}}$ with the budget regularization $R$. Let $\lambda$ denote a hyperparameter controlling the magnitude of the regularization term. The final training loss is a weighted sum of these components:
\begin{align}
\label{eq:loss_function}
   \min_{\theta}  L_{CE}(X,W,S) + \lambda R(b, B(S)) 
\end{align}
\par In this work, we use a similar hypernetwork architecture designed in DISP-LLM, which is described in more detail in Section 5 in the Appendix. 
\par Regarding the training of the hypernetwork, let $o_l$ denote the continuous output produced by the hypernetwork. To obtain the final discrete binary masks $s_l \in \{0,1\}^{d_l}$, we follow a procedure similar to DISP-LLM~\cite{gao2024disp}: we apply the Gumbel-Softmax trick~\cite{jang2016categorical} combined with the Straight-Through Estimator~\cite{bengio2013estimating} to generate differentiable yet discrete binary vectors $s_l$ from the continuous vectors $o_l$. This enables end-to-end gradient propagation through the discrete mask selection during training. Full details of this differentiable binarization step are provided in Section 5.2 of the Appendix.

\par \textbf{Advancement beyond SoTA:} While DISP-LLM targets only structured pruning, our method extends the hypernetwork paradigm to first support MPQ and then enable joint structured pruning and MPQ for LLMs. Additionally, unlike prior layer-wise MPQ methods that consider only local layer-level errors, motivated by DISP-LLM\cite{gao2024disp}, our approach trains the hypernetwork directly on end-to-end language modeling loss. This global perspective yields superior quantization quality compared to the SoTA layer-wise baselines (see Section~\ref{sect:exp_compare_ptq}).
\section{Method}
\subsection{Quantization with binary masks}
\label{subsect:quant_with_binary_mask}
This section outlines the general framework for mixed-precision quantization (MPQ) in LLMs. We define a binary mask $M$ to distinguish between sensitive and non-sensitive weight channels. Specifically, an element $m_i \in \{0, 1\}$ within the mask is set to 1 if the $i$-th channel is identified as salient, and 0 otherwise. Let $Q_{\text{h}}(\cdot)$ and $Q_{\text{l}}(\cdot)$ denote the quantization functions that map weights to high precision (e.g., 4-bit or 8-bit) and low precision (e.g., 1-bit, 2-bit, or 3-bit), respectively. The mixed-precision quantized weight matrix is then formulated as:
\begin{align}\label{eq:quant}
    F_{quant}(W, M) =  Q_{h}(W) \cdot M +  Q_{l}(W)  \cdot (1-M)
\end{align}
\par Similarly, let denote X as the activation input to a linear layer. For activation quantization, we apply the same formulation as follows
\begin{align}
    F_{quant}(X, M) =  Q_{h}(X) \cdot M +  Q_{l}(X)  \cdot (1-M)
\end{align}

Previous MPQ methods \cite{zhao2024atom,huang2024billm,zhao2025ptq1,huang2024slim} derive quantization masks for each layer based on the layer's quantization error. These approaches typically rely on a calibration dataset to evaluate weight importance via activation magnitudes or Hessian-based sensitivity analysis. By applying a greedy search at the layer level, as in previous work, they assume that the quantization choices between layers are independent and overlook cumulative errors and inter-layer dependencies, which leads to sub-optimal solutions. In contrast, we resolve this searching problem by considering the actual final loss of the model in an end-to-end manner, which will be described in more detail in Section \ref{subsect:hypernetwork}.
\par Notably, we reorder the input channels of each weight matrix based on their corresponding activation magnitudes. Following prior work~\cite{zhao2024atom}, these magnitudes are estimated as the mean absolute activation value computed over a calibration set of 128 samples from WikiText-2. This reordering strategy clusters high-magnitude outlier channels into the same groups, preventing a few extreme values from inflating the quantization scales of more sensitive weights. Additionally, consistent with previous MPQ methods~\cite{zhao2024atom,saxena2024resq,liuspinquant}, we adapt GPTQ~\cite{frantar2022gptq} to further refine the quantized weights and reduce residual quantization error. Ablation studies on Llama-2-7B (Table~\ref{tab:ablation_study}) confirm that incorporating both activation-based channel reordering and GPTQ refinement consistently improves the perplexity of the quantized model.




\subsection{Combining quantization and pruning}
\par When combining quantization and pruning, from Equation \ref{eq:prune} and Equation \ref{eq:quant}, we have two possible options, namely, 
\begin{enumerate}
    \item Quantization-then-pruning: $F_{prune}(F_{quant}(W, M), P_{in}, P_{out})$ 
    \item Pruning-then-quantization: $F_{quant}(F_{prune}(W, P_{in}, P_{out}), M)$
\end{enumerate}
\par In literature, structured pruning is generally employed as an initial phase to extract an optimized subnetwork. This architecture is subsequently quantized, yielding a more compact model with improved inference throughput accuracy on target hardware \cite{han2016deep} and achieves better perplexity for LLMs \cite{harmaeffective}. Based on these observations from previous works, we choose the second option for jointly pruning and quantization: the weight matrix is first pruned, and the resulting sparse weights are subsequently quantized. The final joint pruned-then-quantized weight matrix is obtained by combining Equation \ref{eq:quant} and Equation \ref{eq:prune} as follows:
\begin{align}\label{eq:combine}
    F_{quant}(P^T_{in}WP_{out}, M) &= Q_{h}(P^T_{in}WP_{out}) \cdot M +  Q_{l}(P^T_{in}WP_{out})  \cdot (1-M)
\end{align}
\par where $P^T_{in}WP_{out}$ are the weight matrix $W$ pruned by binary vectors for input dimension $P_{in}$ and output dimension $P_{out}$. Similarly, the activation matrix can also be compressed as follows:
\begin{align}\label{eq:combine}
     F_{quant}(XP_{in}, M) &= Q_{h}(XP_{in}) \cdot M +  Q_{l}(XP_{in})  \cdot (1-M)
\end{align}

\par where $XP_{in}$ is the activation matrix $X$ pruned by the binary vector for input dimension $P_{in}$. It is noteworthy that, different from the weight matrix, the activation matrix can only be pruned in the input dimension. 
\par
\begin{figure*}[t]
    \centering
    \includegraphics[width=0.86\textwidth]{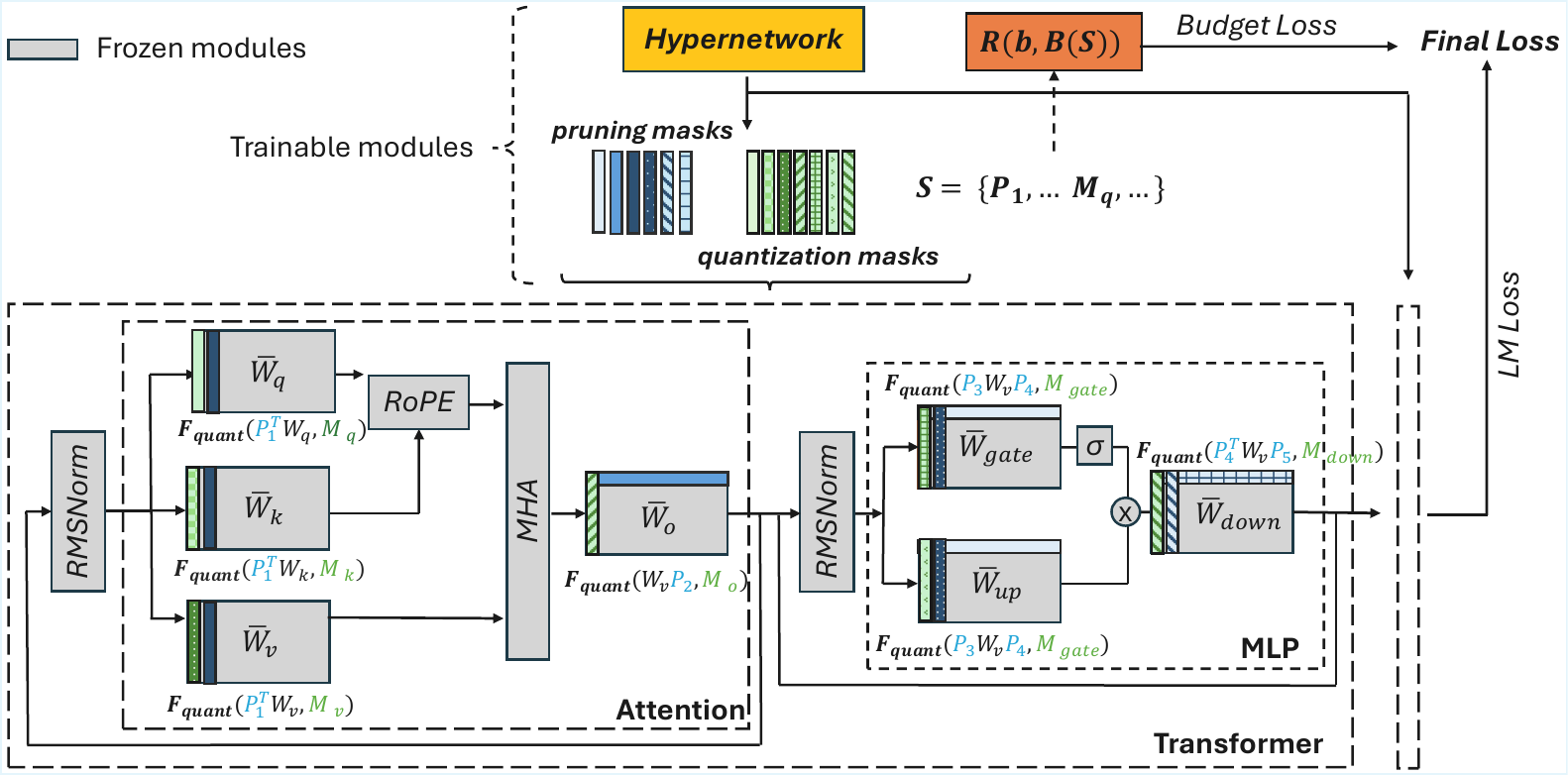}
    \caption{Overview of the masked compressible supernet in \textbf{TOGA}. The gray modules are frozen. Otherwise, colorful modules are trainable.}
    \label{fig:overview}
\end{figure*}

 By applying the pruning and quantization operations (as defined in the above equations) to the weights of linear layers and their corresponding activations in LLMs, we construct a masked compressible supernet. This supernet encodes all possible combinations of structured pruning and MPQ policies within a single model architecture. Unlike conventional sequential approaches that first prune LLMs to a target sparsity level and then quantize the pruned model to a desired precision, our method constructs a supernet that jointly encodes all feasible combinations of structured pruning and mixed-precision quantization policies. The hypernetwork is then trained end-to-end to progressively identify the optimal joint configuration. This integrated search enables mutual awareness during optimization: when determining pruning decisions, the hypernetwork accounts for the current quantization choices, and vice versa. Figure \ref{fig:overview} depicts an overview of the masked compressible supernet proposed in this paper. We name our joint-pruning-and-quantization method as \textbf{TOGA}.

\section{Experiments}
\label{sect:exp}
\subsection{Settings}
\par \textbf{Datasets} We trained our hypernetworks using WikiText-2~\cite{merity2016pointer} dataset. To ensure a fair comparison, other calibration-based methods were evaluated using the same datasets employed for our hypernetwork training.

\par  \textbf{Evaluation Benchmarks}. Our evaluation follows the standard pipeline established in prior literature~\cite{zhao2025ptq1,gao2024disp}. Particularly, we assess model performance using the following metrics and datasets:
\begin{itemize}
    \item Perplexity: Measured using the WikiText-2~\cite{merity2016pointer} and C4~\cite{raffel2020exploring} dataset.
    \item Zero-shot accuracy on Common Reasoning: Directly evaluating 
    pruned models with further fine-tuning steps (zero-shot) on common reasoning benchmarks, including ARC Easy/Challenge~\cite{allenaiarc}, BoolQ~\cite{clark2019boolq}, Winogrande~\cite{ai2_winogrande}, Hellswag~\cite{zellers2019hellaswag}, and  MMLU~\cite{hendrycks2021ethics}. 
\end{itemize}
\par \textbf{Models} We conducted experiments across several widely-used models, including  Llama-3.2-1B, Llama-3.2-3B, Llama-2-7B,  Llama-2-13B, Llama-3-8B, Llama-3.1-8B, Mistral-7B-v0.3 (Mistral-7B), and Qwen-3-8B.

\textbf{Experimental Setup.}
We trained the hypernetworks using a single NVIDIA A100 GPU with 80GB of VRAM. The training duration varied by task: 2,000 steps were allocated for quantization-only experiments, whereas 10,000 steps were used for joint pruning and quantization. All experiments utilized a batch size of 1 and were repeated five times to report the average results. For more  Detailed hyperparameter configurations, see Table 1 of the Appendix.

\par \textbf{Baselines} We compare our proposed method against the following baselines:
\begin{itemize}
\item \textbf{Post-Training Quantization (PTQ) methods}: To isolate the effect of quantization in our approach, we disable pruning by setting all entries in the pruning vectors to ones. We refer to this quantization-only variant as \textbf{TOGA-q}. In this quantization-only setting, the desired budget $b$ is defined as the percentage of salient weights. We then compare \textbf{TOGA-q} against SoTA PTQ techniques, including:
\begin{itemize}
    \item Weight-activation PTQ: We compare against SOTA mixed-precision methods, namely Atom~\cite{zhao2024atom} and ResQ~\cite{saxena2024resq}, as well as a strong uniform-precision baseline, SpinQuant~\cite{liuspinquant}. Notably, all weight-activation quantization baselines, including our proposed \textbf{TOGA-q}, employ the GPTQ quantizer~\cite{frantar2022gptq} to further improve the quality of the quantized models. For GPTQ, we use a group size of 64 for Llama-3.2-1B and a group size of 128 for all other models.
    \item Weight-only PTQ: We include PTQ-1.61~\cite{zhao2025ptq1} (evaluated both with and without LoRA preprocessing), SliM-LLM~\cite{huang2024slim}, and BiLLM~\cite{huang2024billm}. For 1-bit quantization, we adopt a binarization approach consistent with that used in PTQ-1.61~\cite{zhao2025ptq1}.
\end{itemize}
\item \textbf{Sequential pruning + quantization:} We compare \textbf{TOGA} (joint pruning and quantization) against a sequential baseline: DISP-LLM pruning followed by PTQ (Atom, ResQ, BiLLM, SliM-LLM). For a fair comparison, we also evaluate a variant of \textbf{TOGA} by adding a sparsity constraint to the loss function in Equation \ref{eq:loss_function}, and denote it as \textbf{TOGA-fixed-sparsity}. We also include two strong joint baselines: SparseGPT+GPTQ~\cite{frantar2023sparsegpt} and OBR~\cite{guo2025optimal}. In this joint pruning-and-quantization setting, the desired budget $b$ is compression budget defined in Section \ref{subsect:define}.
\end{itemize}

\subsection{Definitions}
\label{subsect:define}
For clarity and consistency in the experiments, we define the following key terms used throughout the paper.
\par \textbf{Compression Budget:} The compression budget (or compression ratio) is defined as the ratio of the theoretical memory footprint of the compressed model to that of the original FP16 model. For instance, 50\% sparsity combined with W4A4 quantization gives a budget of
$0.5 \times \frac{4}{16} = 0.125$ (i.e., the model retains 12.5\% of the original memory, or achieves 87.5\% memory reduction).

\par  \textbf{Precision Format:} In mixed-precision quantization, the effective average bit-width depends on the chosen precisions for salient and non-salient portions and their respective proportions. For instance, Atom quantizes 97\% of weight and activation channels to 3 bits while preserving 3\% of salient weights at 8 bits, giving an average of $\approx 3.2$ bits (denoted W(3.2)A(3.2)). Likewise, PTQ-1.61 quantizes 80\% of weights to 1 bit and 20\% to 4 bits (activations remain 16-bit), yielding an average weight bit-width of 1.6 (denoted W(1.6)A16).

\subsection{Comparison with other mixed-precision PTQ methods}
\label{sect:exp_compare_ptq}
\par We first evaluate \textbf{TOGA-q} against SoTA PTQ methods at ultra-low precisions (sub-2-bit weight-only and sub-4-bit weight-activation) before comparing all approaches at the more hardware-friendly INT4 and INT8 settings.
\par \textbf{Ultra-low precision:} We compare \textbf{TOGA-q} with several SoTA mixed-precision PTQ methods. Atom~\cite{zhao2024atom} preserves $\sim$3\% of salient weights (the last 128 channels of each weight matrix) at 8-bit precision, yielding an average bit-width of 3.2 bits. Similarly, we adopt the same setup for ResQ to ensure fairness. For \textbf{TOGA-q}, we match this budget by assigning the desired budget $b$ to 3\%, and quantizing salient weights to 8-bit and the rest to 3-bit, directly following the Atom configuration. In the weight-only setting, BiLLM achieves a theoretical average of 1.1 bits but requires additional unstructured binary masks to group salient weights, incurring non-negligible metadata overhead and resulting in an effective average closer to 2.1 bits~\cite{zhao2025ptq1}. In contrast, both PTQ-1.61 and \textbf{TOGA-q} allocate 80\% of each weight matrix to 1-bit and 20\% to 4-bit, with virtually no metadata overhead; this exact combination reproduces the PTQ-1.61 results faithfully.
\par Table~\ref{tab:mix_quant} presents perplexity results on WikiText-2 along with the average zero-shot accuracy across six standard reasoning benchmarks for the quantized models. Further results for weight-only PTQ methods are provided in Appendix Section 3.1. Overall, \textbf{TOGA-q} consistently surpasses all other SoTA approaches evaluated. Compared to ResQ, the strongest baseline among weight-activation PTQ methods, \textbf{TOGA-q} reduces WikiText-2 perplexity by 5–21\% and improves average zero-shot accuracy on the reasoning tasks by 1.6–4.5\%. In particular, the leading uniform-precision method SpinQuant experiences total perplexity collapse at 3-bit precision, underscoring the importance of mixed-precision quantization to maintain model coherence in sub-4-bit settings by safeguarding critical weights.
\par Additionally, although PTQ-1.61$^\dagger$, one of the most competitive weight-only PTQ baselines, relies on a costly preprocessing step (10,000 LoRA fine-tuning iterations on RedPajama), \textbf{TOGA-q} requires neither external datasets nor any fine-tuning, yet it achieves 9–59\% lower perplexity on WikiText and 2.3–5.4\% higher average zero-shot accuracy on the reasoning tasks. While SliM-LLM remains reasonably competitive, its approach of using three different bit-widths per linear layer introduces extra kernel-launch overhead during inference, frequently resulting in no speedup, or even performance regression, compared to the FP16 baseline~\cite{huang2024slim}. By contrast, \textbf{TOGA-q} uses only two bit-widths per layer, delivers 3–50\% lower perplexity on WikiText-2, 16–85\% lower perplexity on C4 (see Tables 3 and 4 in the Appendix), and achieves 2-4.21\% higher average zero-shot accuracy in the six reasoning tasks.

\begin{table*}[t] \scriptsize
\caption{Perplexities (lower is better) on WikiText-2 dataset and average accuracy (higher is better) of six different reasoning tasks of models quantized by different mixed-precision PTQ methods.  (PTQ-1.61$^\dagger$ means that we apply PTQ-1.61 with a pre-processing step using RedPajama dataset)}
\label{tab:mix_quant}
\centering
\begin{tabular}{c c |c c c c | c  c c c c}
\hline
\multirow{2}{*}{Method} &  \multirow{2}{*}{Precision} & \multicolumn{4}{c|}{Perplexity ($\downarrow$)} & \multicolumn{4}{c}{Zero-shot Accuracy ($\uparrow$)} \\
\cline{3-10}
  & & 2-7 & 3-8 &  \multirow{1}{*}{Mistral-7B} & \multirow{1}{*}{Qwen3-8B} & 2-7 &3-8  &  \multirow{1}{*}{Mistral-7B} &  \multirow{1}{*}{Qwen3-8B}\\

\hline
\makecell[c]{PTQ-1.61} & W(1.6)A16 & 22.65 &  805.63 & 45.69 & 160.02 & 34.22 & 33.76 & 37.15 &   34.02\\
\multirow{1}{*}{BiLLM} & W(2.1)A16 & 32.58 & 36.01 & 29.90 & 47.32 & 41.16 & 39.16 & 39.11 & 37.87   \\
\multirow{1}{*}{SilM-LLM} & W2A16 & 16.01 & 40.60 & 16.37 &  23.93 & 43.94 & 36.12 &  40.03 & 43.00 \\
PTQ-1.61$^\dagger$ & W(1.6)A16 & 12.70 & 22.35 & 38.51 & 28.34 & 41.43 & 35.59  & 39.77 & 43.53 \\
\multirow{1}{*}{\textbf{TOGA-q}} & W(1.6)A16 & \textbf{11.00} & \textbf{20.28} & \textbf{15.93} & \textbf{20.29} & \textbf{46.80} & \textbf{41.33} & \textbf{42.07}  & \textbf{47.58} \\
\hline
SpinQuant & W3A3 & 438.11 & 205.43 & 21.63 & - & 33.06 & 34.26 & 33.96  & -\\
Atom & W(3.2)A(3.2) & 12.12 & 48.53 & 10.34 & 38.78 & 48.13 & 39.57 & 51.71 &  38.36 \\
ResQ & W(3.2)A(3.2) & 7.35 & 15.53 & 6.53 & 15.60 & 52.63 & 46.33 & 58.43 & 42.45  \\
\textbf{TOGA-q} & W(3.2)A(3.2) & \textbf{7.30} & \textbf{12.24} & \textbf{6.40} & \textbf{14.16} & \textbf{54.20} & \textbf{50.81} & \textbf{62.60} & \textbf{56.03} \\
\hline
\end{tabular}
\end{table*}
\par \textbf{INT4+INT8 Mixed Precision}: In this experiment, we employ a combination of INT4 and INT8 precisions for mixed-precision PTQ methods, as both precisions are hardware-friendly and natively supported by NVIDIA GPUs. We evaluate our \textbf{TOGA-q} method against leading mixed-precision PTQ approaches, specifically Atom and ResQ. To further demonstrate the advantages of mixed-precision quantization, we also include SpinQuant, a strong representative of SoTA uniform-precision PTQ methods. For the mixed-precision PTQ baselines, we adopt the same configuration as ResQ: 12.5\% of the weights are identified as salient and quantized to INT8, while the remaining weights are quantized to INT4. The same setting is also applied to Atom in this experiment. Additionally, since the KV cache constitutes one of the largest contributors to memory usage during LLM inference and deployment, we further reduce its memory footprint by quantizing the KV cache to INT4 precision. Table~\ref{tab:mix_quant_reasoning_w4a4kv4} reports the average zero-shot accuracy of Llama-family models quantized via these PTQ methods on six reasoning tasks. Overall, consistent with trends observed in ultra-low-bit settings, all MPQ methods outperform SpinQuant, the leading uniform-precision baseline. More importantly, \textbf{TOGA-q} surpasses Atom, which simply identifies and preserves the top 12.5\% most sensitive activation channels (along with their corresponding weight channels) at higher precision while aggressively quantizing the rest. This result underscores the effectiveness of \textbf{TOGA-q}'s adaptive, loss-driven bit-width allocation across layers. Finally, \textbf{TOGA-q} also outperforms ResQ, the strongest prior MPQ baseline.

\begin{table*}[t] \scriptsize
\caption{Zero-shot accuracy (higher is better) of the Llama-family models under different post-training quantization (PTQ) methods across various reasoning tasks. SpinQuant applies uniform W4A4 quantization. All mixed-precision methods quantize 12.5\% of the weights (salient weights) to INT8, with the remaining weights quantized to INT4. For all methods, the KV cache is quantized to INT4 precision.}
\label{tab:mix_quant_reasoning_w4a4kv4}
\centering
\begin{tabular}{c | c c c c c c c c }
\hline
 \multirow{2}{*}{Model} & \multirow{2}{*}{Method} & \multicolumn{6}{c}{Zero-shot Task Performance ($\uparrow$)} & \multirow{2}{*}{Average} \\
\cline{3-8}
& & ARC\_e & ARC\_c & BoolQ & Hellaswag & WinoGrande & MMLU\\
\hline

\multirow{5}{*}{Llama-3.2-1B} & 16-bit baseline & 60.6 & 36.5 & 63.4 &  63.6 & 60.1 &  32.8 & 52.8 \\ 
\cdashline{2-9}
 & SpinQuant &  51.8 & 32.3 & 59.3 & 55.4 &  54.7 & 24.9 & 46.4 \\
& Atom & 58.8 & 33.7 & 58.9 & 56.7 & 53.0 & \textbf{28.6} & 48.3 \\
& ResQ & 56.6 & 33.6 & 58.9 & \textbf{58.2} & 55.3 & 26.6 & 48.2 \\
& \textbf{TOGA-q} & \textbf{59.7} & \textbf{33.8} & \textbf{59.9} & 57.7 & \textbf{57.4} & 26.8 & \textbf{49.2} \\
\hline 
\multirow{5}{*}{Llama-3.2-3B} & 16-bit baseline &  71.7 &  46.2 & 73.1 & 73.7 & 69.1 &  49.5 & 63.9  \\ 
\cdashline{2-9}
 & SpinQuant &  64.8 & 38.9 & 68.0 & 69.1 & 62.9 & 37.2 & 56.8 \\
& Atom & 69.0 & 40.0 & 64.2 & 69.1 & 62.4 & 44.0 & 58.1 \\
& ResQ & 65.6 & \textbf{43.1} & 68.8 & \textbf{70.5} & \textbf{64.8} & \textbf{47.8} & 60.1 \\
& \textbf{TOGA-q} & \textbf{70.5} & 41.1 & \textbf{72.6} & 70.2 & 63.2 & 46.5 & \textbf{60.7} \\
\hline 
\multirow{5}{*}{Llama-2-7B} & 16-bit baseline & 74.6 & 46.3 & 77.8 &  75.9 & 69.1 &  39.5 & 63.9 \\ 
\cdashline{2-9}
 & SpinQuant & 71.3 & \textbf{43.6} & 73.8 &  73.2 & 65.4 & 33.5 & 60.1 \\
& Atom & 74.0 & 42.7 & 74.9 & 73.5 & 66.9 & 34.8 & 61.1 \\
& ResQ & 72.0 & 43.3 & \textbf{75.9} &  73.9 & 66.8 & 37.3 & 61.5 \\
& \textbf{TOGA-q} & \textbf{74.5} & 43.5 & 75.6 & \textbf{74.3} & \textbf{69.3} & \textbf{37.6} & \textbf{62.5}  \\
\hline 
\multirow{5}{*}{Llama-3-8B} & 16-bit baseline &  77.1 &  53.2 & 81.1 & 79.2 & 73.4 & 56.7 & 70.1 \\ 
\cdashline{2-9}
 & SpinQuant & 75.4 &  48.0 & 75.8 & 75.4 &  69.2 & 51.2 & 65.8 \\
& Atom & 75.9 & 47.4 & 73.5 & 75.5 & 68.6 & 54.0 & 65.8 \\
& ResQ & 75.0 & \textbf{49.2} & 72.5 & 76.5 & \textbf{71.0} & \textbf{56.6} & 66.8 \\
& \textbf{TOGA-q} & \textbf{76.3} & 47.1 & \textbf{76.9} & \textbf{76.8} & 70.1 & 56.3 &  \textbf{67.2} \\
\hline 
\end{tabular}

\end{table*}

\subsection{Comparison with other pruning+quantization methods}
\label{subsection:prune_quant}
\par
\begin{figure*}[htbp]
     \centering
     \begin{subfigure}[b]{0.44\textwidth}
         \centering
         \includegraphics[width=\textwidth]{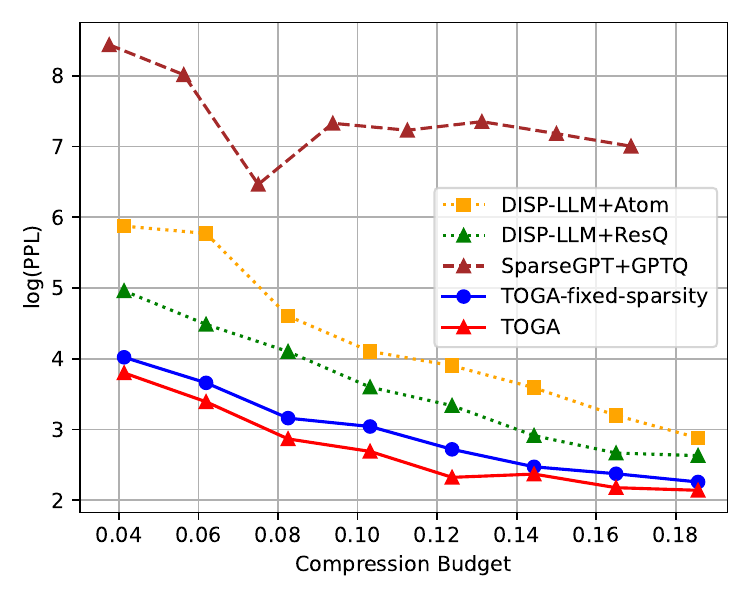}
         \caption{Weight-Activation}
         \label{fig:plot1}
     \end{subfigure}
     \hfill
     \begin{subfigure}[b]{0.44\textwidth}
         \centering
         \includegraphics[width=\textwidth]{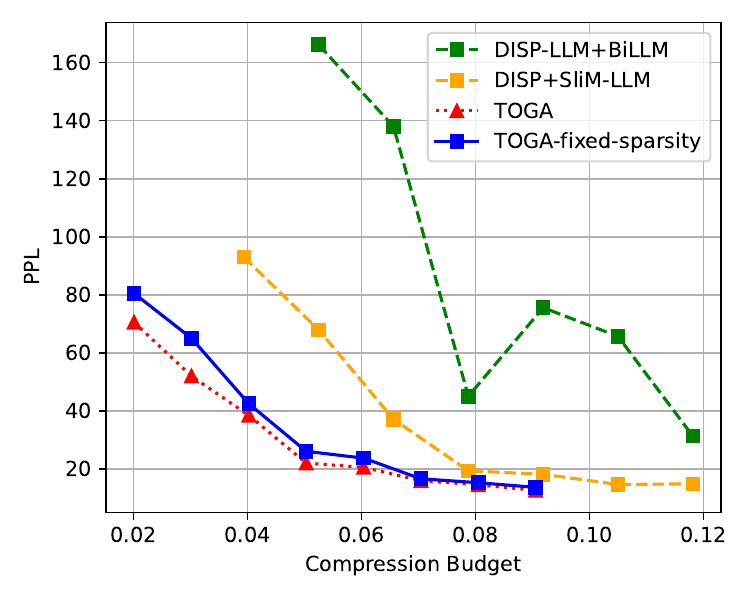}
         \caption{Weight-Only}
         \label{fig:plot3}
     \end{subfigure}
     
     \caption{Perplexity on WikiText2 dataset of Llama-2-7B compressed by weight-activation (left) and weight-only (right) quantization methods at different compression budgets. (Left: all methods are quantized to W(3.2)A(3.2) format. Right: DISP+BiLLM is quantized to W(2.1)A16 format as explained in~\cite{zhao2025ptq1}. Other methods are quantized to W(1.6)A16 format.)}
     \label{fig:prune+quant}
\end{figure*}

 In this section, we compare TOGA with other joint pruning and quantization baselines. Figure~\ref{fig:prune+quant} compares our joint structured pruning and mixed-precision quantization method (\textbf{TOGA}) against sequential baselines: DISP-LLM structured pruning (sparsity 0.2--0.9) followed by PTQ via Atom, ResQ, BiLLM, or SliM-LLM. In contrast, \textbf{TOGA} jointly optimizes both pruning and quantization in a single unified framework. This joint optimization allows \textbf{TOGA} to flexibly balance sparsity and precision for any target compression ratio (e.g., at $\sim$0.103, it can yield either 45\% sparsity + W3A3 or 59\% sparsity + W4A4). For fair comparison, we add a regularization term (Equation~\ref{eq:loss_function}) to constrain \textbf{TOGA} to the exact sparsity levels of the baselines; we refer to this variant as \textbf{TOGA-fixed-sparsity}. Furthermore, we exclude OBR due to its failure at ultra-low precision (e.g., Llama-2-7B has perplexity $>1500$ at 10\% sparsity with W3A3 precision, or 222.67 with W2A16 precision); further comparison with the OBR baseline at INT4  precision is included in Table 7 in the Appendix.
\par The left panel of Figure~\ref{fig:prune+quant} shows weight-activation PTQ results at effective W(3.2)A(3.2) across compression ratios 0.04 (80\% sparsity) to 0.18 (10\% sparsity), including the unstructured SparseGPT+GPTQ baseline (pruned then quantized to W3A3). The right panel evaluates weight-only PTQ baselines. We omit PTQ-1.61 (requires costly LoRA fine-tuning on an external dataset) and SparseGPT+GPTQ (catastrophic failure at W2A16, e.g., perplexity $>7000$ for Llama-2-7B at 10\% sparsity).
\par Overall, \textbf{TOGA} consistently outperforms both SoTA joint and sequential (DISP-LLM + other PTQ methods) pipelines. Unconstrained \textbf{TOGA} surpasses even \textbf{TOGA-fixed-sparsity}, particularly at aggressive compression levels, confirming that simultaneous optimization and free sparsity-precision trade-offs deliver superior accuracy-efficiency Pareto fronts. Please refer to Appendix Section 3.2 for more experiments.
    
\subsection{Performance}
\par In this experiment, we evaluate the practical performance gains of \textbf{TOGA} relative to OBR and FP16 baselines using the Llama-2-7B model. For our \textbf{TOGA} method, we developed custom CUDA kernels based on the CUTLASS library to support mixed-precision INT4/INT8 GEMM operations. We further optimized the inference pipeline by fusing the RMS Normalization layer with reordering and quantization stages to minimize computational overhead (see Section 4 in the Appendix for implementation details). 
Besides that, we also quantize the KV cache to reduce the memory overhead during the inference phase.
\par
\begin{figure*}[t]
     \centering
         \centering
         \includegraphics[width=0.85\textwidth]{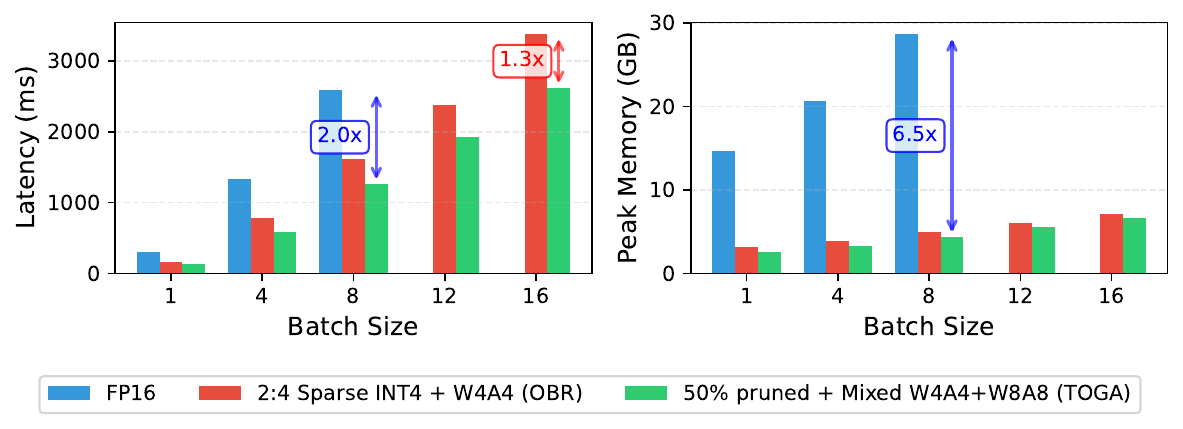}
     \caption{Inference latency during the prefill stage and peak memory usage during the decode stage for the Llama-2-7B model, compressed using baselines and \textbf{TOGA}. All measurements were conducted with a fixed context length of 2048 tokens and varying batch sizes. The FP16 baseline encounters out-of-memory errors at batch sizes $\geq$ 12.}
     \label{fig:speed_memory}
\end{figure*}

 For OBR method, to ensure a fair and reproducible comparison, we use CUTLASS-based kernels from the OBR paper that support INT4 2:4 semi-structured sparse GEMM operations. These kernels were integrated into the QuaRot inference pipeline \cite{ashkboos2024quarot}, similar to that in the original OBR paper \cite{guo2025optimal}. It is noteworthy that OBR also quantizes the KV cache to INT4. For the FP16 baseline, we use the standard HuggingFace model without any modifications. All performance benchmarks were performed on a NVIDIA L40 GPU, with results averaged over 100 runs using a fixed context length of 2048 tokens and varying batch sizes. The performance evaluation follows a similar protocol to that described in \cite{ashkboos2024quarot} and is summarized as follows:
\begin{itemize}
    \item \textbf{Compute-bound prefill stage:} The left side of Figure \ref{fig:speed_memory} shows inference speedups for OBR and \textbf{TOGA} across different batch sizes at a context length of 2048. Notably, the FP16 baseline encounters out-of-memory (OOM) errors starting at batch size 12. In contrast, OBR and \textbf{TOGA} can sustain significantly higher batch sizes without memory issues. Moreover, \textbf{TOGA} delivers superior prefill speedups compared to the 2:4 semi-structured OBR baseline. In particular, \textbf{TOGA} achieves up to $2\times$ speedup over the FP16 baseline and approximately $1.3\times$ speedup over OBR baseline in the prefill phase.
    \item \textbf{Memory-bound decoding phase:} Since memory consumption is the primary bottleneck during autoregressive decoding, we focus on peak memory usage. \textbf{TOGA} reduces peak memory by up to $6.5\times$ compared to the FP16 baseline and by approximately 10\% compared to OBR baseline.
\end{itemize}
\par These results demonstrate the substantial practical advantages of \textbf{TOGA}’s joint structured pruning and mixed-precision quantization approach over both the uncompressed FP16 baseline and the 2:4 semi-structured sparsity method employed by OBR. For accuracy on reasoning tasks and perplexity evaluation of models compressed by \textbf{TOGA} and OBR, please see Table 7 in the Appendix.

\subsection{Ablation Study}
\textbf{Quantization techniques with TOGA-q:} In this experiment, we first use round-to-nearest (RTN) to adopt per-channel quantization for weights and per-token quantization for activations, which is also the standard quantization recipe \cite{xiao2023smoothquant}, to uniformly quantize the model to W4A4. Then, we apply quantization techniques used in \textbf{TOGA-q}, namely quantizing 12.5\% of weights to INT8, reordering weight channels (see Section \ref{subsect:quant_with_binary_mask}), GPTQ~\cite{frantar2022gptq}, and quantizing the KV cache to INT4. As shown in Table~\ref{tab:ablation_study} for Llama-2-7B on WikiText-2 and C4, retaining 12.5\% of sensitive weights at INT8 already yields a large perplexity reduction compared to uniform W4A4. Adding channel reordering and GPTQ further improves performance from 6.03 to 5.38, while INT4 KV-cache quantization causes only a small increase to 5.48, demonstrating that the overall method remains highly effective even under KV cache quantization.

\begin{table}[t]\scriptsize
\centering
\caption{Ablation Study on quantization techniques with \textbf{TOGA-q} for Llama-2-7B model to mixture precision of W4A4 and W8A8. We maintain an average of 12.5\% of rows of weight and activation matrices in W8A8 format.}
\label{tab:ablation_study}
\begin{tabular}{|l|c c|}
\hline
\multirow{2}{*}{\textbf{Quantization Techniques}} &   \multicolumn{2}{c|}{\textbf{Perplexity}}  \\
\cline{2-3}
& WikiText-2 & C4 \\
\hline

16-bit baseline & 5.12& 7.10 \\
\cdashline{1-3}
W4A4 RTN & 1753 & 2301 \\
+ Quantizing 12.5\% of weights to INT8 & 6.03 & 8.10 \\
+ Reordering & 5.78 & 8.01 \\
+ GPTQ &  \textbf{5.38} & \textbf{7.47} \\
+ Quantizing KV cache to INT4 &  5.48 & 7.68 \\
\hline

\end{tabular}
\end{table}

\textbf{Salient Weight Distribution} In this experiment, we investigate the quantization choices made by \textbf{TOGA-q}. In contrast to previous mixed-precision methods such as Atom and ResQ, which apply a uniform, fixed fraction of salient weights (quantized to INT8) across all layers while quantizing the rest to INT4, \textbf{TOGA-q} adopts a more flexible and adaptive approach. It utilizes a hypernetwork to automatically search for and identify the optimal number of salient weights to be kept at INT8 in each individual layer, with the remainder quantized to INT4. This enables layer-specific mixed-precision assignments that more effectively account for the varying redundancy and sensitivity across different layers.
\par
\begin{figure*}[t]
     \centering
     \begin{subfigure}[b]{1\textwidth}
         \centering
         \includegraphics[width=0.9\textwidth]{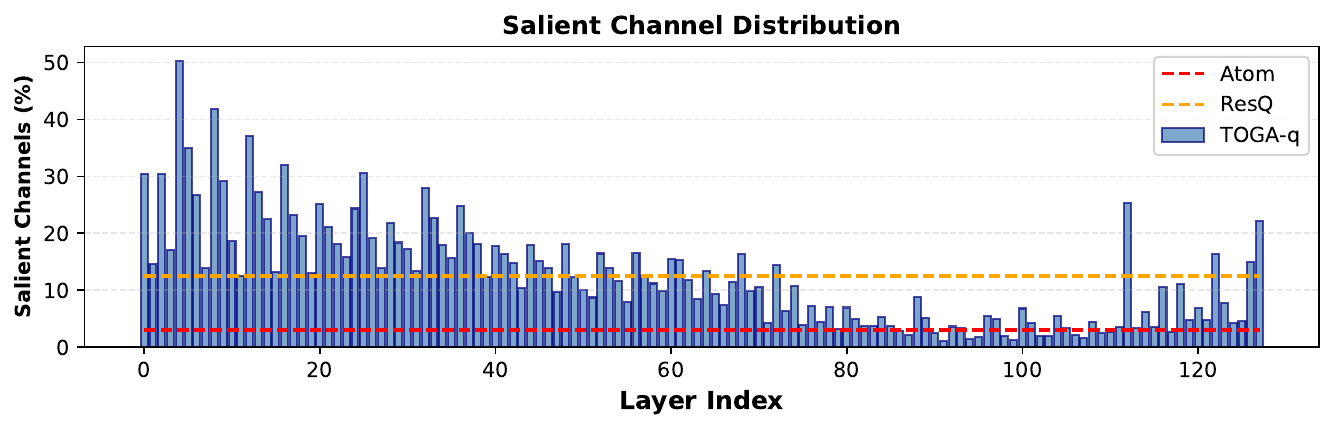}
         \label{fig:plot1}
     \end{subfigure}
     
     \caption{Distribution of Salient Channels suggested by \textbf{TOGA-q}, Atom, and ResQ when quantizing  non-salient/salient weights to INT4/INT8 for Llama-2-7B.}
     \label{fig:salient_weights}
\end{figure*}

 Figure \ref{fig:salient_weights} illustrates the distribution of salient channels  as determined by \textbf{TOGA-q} and other baselines on Llama-2-7B. Notably, \textbf{TOGA-q} tends to allocate a larger number of salient weights to the first 16 transformer blocks and to a few of the final blocks, while assigning significantly fewer salient weights to the intermediate blocks. This pattern aligns well with existing empirical findings on layer importance in large language models \cite{sreenivas2024llm}, which show that removing the early layers or the last few layers causes substantially larger accuracy degradation compared to pruning the middle layers. In contrast, ResQ and Atom apply the same threshold for all linear layers to identify salient weights.

\section{Conclusions}
\vspace{-0.1cm}
In this work, we present a novel mixed-precision post-training quantization (PTQ) method coupled with the first end-to-end framework for joint structured pruning and quantization of LLMs. By optimizing bit-width allocation via a hypernetwork trained on end-to-end language modeling loss and integrating it with structural pruning in a unified search space, our method achieves substantial inference acceleration over the uncompressed FP16 baseline, while delivering superior perplexity on language modeling datasets, higher zero-shot accuracy on downstream reasoning benchmarks, and improved real-world throughput (prefill and decode phases) compared to SoTA approaches that combine semi-structured pruning with quantization. These gains enable practical deployment of LLMs on severely resource-constrained hardware.
\section{Future Works}
A major limitation of our proposed method is that it requires loading and running the entire LLM on GPUs, which can easily lead to Out-of-Memory (OOM) errors with very large models (e.g., 70B parameters). For instance, under the current implementation, a GPU with 80GB of memory can only accommodate up to a 32B-parameter LLM. In the future, we plan to leverage advanced distributed training or offloading techniques to significantly reduce memory consumption during hypernetwork training.


%
\bibliographystyle{splncs04}
\bibliography{ecml}
\end{document}